\documentclass[letterpaper, 10 pt, conference]{ieeeconf}  

\IEEEoverridecommandlockouts                              

\usepackage[utf8]{inputenc} 
\usepackage[T1]{fontenc}    
\usepackage{hyperref}       
\usepackage{url}            
\usepackage{booktabs}       
\usepackage{amsfonts}       
\usepackage{nicefrac}       
\usepackage{microtype}      
\usepackage{xcolor}         
\usepackage{todonotes}

\usepackage{caption}
\usepackage{subcaption}
\usepackage{amsmath}
\usepackage{graphicx}

\usepackage[nolist]{acronym}

\title{A Neural-Evolutionary Algorithm for Autonomous Transit Network Design}

\author{
Andrew Holliday and Gregory Dudek$^{1}$
\thanks{$^{1}$Both authors are with the School of Computer Science, McGill University.  They can be reached at \texttt{ahollid@cim.mcgill.ca} and \texttt{dudek@cim.mcgill.ca}, respectively.}
}

\begin{document}

\maketitle
\thispagestyle{empty}
\pagestyle{empty}

\begin{abstract}
  Planning a public transit network is a challenging optimization problem, but essential in order to realize the benefits of autonomous buses.  We propose a novel algorithm for planning networks of routes for autonomous buses.  We first train a graph neural net model as a policy for constructing route networks, and then use the policy as one of several mutation operators in a evolutionary algorithm.  We evaluate this algorithm on a standard set of benchmarks for transit network design, and find that it outperforms the learned policy alone by up to 20\% and a plain evolutionary algorithm approach by up to 53\% on realistic benchmark instances.

\end{abstract}

\section{Introduction}

Mass public transit is key to sustainable urban design.  Mass transit can reduce traffic congestion, thus shortening trips, and consume less energy, thus reducing emissions, versus even idealized point-to-point transit~\cite{roughgarden2002bad, oh2020evaluating, rich2023fixed}.  Self-driving vehicle technology is bringing with it autonomous bus fleets, which are already being deployed~\cite{malagaAutonomousBus, smartcitiesdriveDetroit, bloombergJacksonville}.  By dispensing with drivers, autonomous buses can enable fleets of more and smaller vehicles~\cite[chapter 9]{cederBook}.  This can make service more frequent and reliable, two major factors in attracting riders to public transit~\cite{Chakrabarti2017getPeopleOutOfTheirCars}.  But frequent service only helps those living near stops; and the need to transfer between lines makes transit less attractive~\cite{Chakrabarti2017getPeopleOutOfTheirCars, el2014newEvidence}.  An efficient network also saves operating costs, which can then be spent on automating and expanding fleets.  So in order to fully realize the benefits of autonomous buses, we need well-designed transit networks.


But the \ac{NDP} is very challenging.  It is similar to the \ac{TSP} and \ac{VRP}, but is much more complex because of its many-to-many nature, and because of transfers between transit lines.  Furthermore, real-world instances of this problem may have hundreds or thousands of transit stops~\cite{montrealTransit}.  The most successful approaches to date have been approximate solvers based on metaheuristics such as \acp{EA}.  These work by repeatedly applying one or more low-level heuristics that randomly modify a solution, and guiding this random search towards more promising solutions over many iterations.  We here wish to consider whether a neural net could learn to use information about the scenario to select promising changes - instead of purely random changes, as existing low-level heuristics do - and whether this could improve the performance of metaheuristic algorithms.

To this end, we train a \ac{GNN} policy via \ac{RL} to construct transit networks for cities.  We then use
this policy as a low-level heuristic in a \ac{EA}.  We compare our approach to the \ac{GNN} model on its own and to the baseline \acl{EA}, and find that on realistically-sized problem instances, the hybrid algorithm performs best, and is competitive with other state-of-the-art approaches.

\section{Related Work}\label{sec:relatedWorks}

\subsection{Graph Nets and Reinforcement Learning for Optimization Problems}

\Acfp{GNN} are neural net models that are designed to operate on graph-structured data \cite{bruna2013spectral,kipf2016semi,defferrard2016spectral,duvenaud2015convolutional}. They were inspired by the success of convolutional neural nets on computer vision tasks and have been applied in many domains, including analyzing large web graphs~\cite{ying2018webscale}, designing printed circuit boards~\cite{mirhoseini2021graph}, and predicting chemical properties of molecules~\cite{duvenaud2015convolutional, gilmer2017quantum}. An overview of \acp{GNN} is provided by~\cite{battaglia2018relational}.

There is growing interest in the application of machine learning techniques to solve \ac{CO} problems such as the \ac{TSP}~\cite{bengio2021machine}.  In \ac{CO} problems generally, it is difficult to find a globally optimal solution but easy to compute a scalar quality metric for a given solution.  As noted by~\cite{bengio2021machine}, this makes \ac{RL}, in which a system is trained to maximize a scalar reward, a good fit.  Many \ac{CO} problems have natural interpretations as graphs, making \acp{GNN} also a good fit.  \cite{vinyals2015pointer} proposed a \ac{GNN} model called a Pointer Network, and trained it by supervised learning to solve \ac{TSP} instances.  Much recent work has built on this by using various \ac{RL} algorithms to train \ac{GNN} to construct \ac{CO} solutions~\cite{dai2017learningCombinatorial, Kool2019AttentionLT, lu2019learning, sykora2020multi}.  These have attained impressive performance on the \ac{TSP}, the \ac{VRP}, and related problems.  \cite{choo2022simulation} use a hybrid of Monte Carlo Tree Search and Beam Search to guide the construction of \ac{CVRP} solutions by a neural net policy.  \cite{mundhenk2021symbolic} trains an \ac{RNN} via \ac{RL} to provide a starting population of solutions to a \ac{EA}, the outputs of which are used to further train the \ac{RNN}.  \cite{fu2021generalize} train a model on small \ac{TSP} instances and propose an algorithm to apply it to much larger instances.  These approaches belong to the family of ``construction'' methods, which solve a \ac{CO} problem by starting with an ``empty'' solution and adding to it in steps, ending once the solution is complete - for example, when all nodes have been visited in the \ac{TSP}.  The solutions from these neural construction methods come close to the quality of those from specialized \ac{TSP} algorithms such as Concorde~\cite{concordeTspSolver}, while requiring much less run-time to compute~\cite{Kool2019AttentionLT}.  

``Improvement'' methods, by contrast with construction methods, start with a complete solution and repeatedly modify it, searching through the solution space for improvements.  Metaheuristics like \ac{EA} belong to this category.  These are more computationally costly than construction methods but can yield better solutions to \ac{CO} problems.  Some work has considered training neural nets to choose the search moves to be made at each step of an improvement method~\cite{hottung2019neural, chen2019learning, d2020learning, wu2021learning, ma2021learning}, and \cite{kim2021learning} train one \ac{GNN} to construct a set of initial solutions, and another to modify and improve them.  This work has shown impressive performance on classic \ac{CO} problems like the \ac{TSP}.  Our approach belongs to this family, but is novel in that it uses an \ac{RL}-trained \ac{GNN} to perform mutations in a \ac{EA}.

\subsection{Optimization of Public Transit}

The \acl{NDP} is NP-complete~\cite{quak2003bus}, making it impractical to solve optimally.  While analytical optimization and mathematical programming methods have been successful on small instances~\cite{vannes2003AnalyticRouteAndSchedule, guan2006AnalyticRoutePlanning}, they struggle to realistically represent the problem~\cite{guihaire2008transitReview, kepaptsoglou2009transitReview}, and so metaheuristic approaches (as defined by~\cite{sorensen2018history}) have been more widely applied.  Historically, \acp{GA}, simulated annealing, and ant-colony optimization have been most popular, along with hybrids of these methods~\cite{guihaire2008transitReview, kepaptsoglou2009transitReview, nikolic2013transit}.

While neural nets have seen much use for predictive problems in urban mobility~\cite{xiong1992transportation, rodrigueNNsForLandUseAndTransport, chien2002dynamic, jeong2004bus, akgungorNNsForAccidentPrediction, li2020graph} and for transit optimization problems such as scheduling and passenger flow control~\cite{zou2006lightrail, ai2022deep, Yan2023DistributedMD, jiang2018passengerInflow}, little work has applied \ac{RL} or neural nets to the \ac{NDP}. ~\cite{darwish2020optimising} and \cite{yoo2023reinforcement} both use \ac{RL} to design a network and schedule for the Mandl benchmark~\cite{mandl1980evaluation}, a single small graph with just 15 nodes. \cite{darwish2020optimising} use a \ac{GNN} approach inspired by~\cite{Kool2019AttentionLT}; in our own work we experimented with a nearly identical approach to~\cite{darwish2020optimising}, but found it did not scale beyond very small instances.  Meanwhile, \cite{yoo2023reinforcement} uses tabular \ac{RL}, a type of approach which also scales poorly.  Both these approaches require a new model to be trained on each problem instance.  Our approach, by contrast, is able to find good solutions for realistically-sized \ac{NDP} instances of more than 100 nodes, and can be applied to problem instances unseen during training.

In prior work~\cite{holliday2023augmenting}, we trained a \ac{GNN} construction method to plan a transit network, and used it to provide initial solutions to the \ac{EA} of~\cite{nikolic2013transit}, finding that this improved the quality of the solutions found.  The present work considers the same \ac{GNN} architecture and training procedure, but uses the \ac{GNN} to generate new routes throughout an improvement method, instead of just once at the outset.

\section{The Transit Network Design Problem}

In the \ac{NDP}, one is given an augmented graph that represents a city:
\begin{equation}
 \mathcal{C} = (\mathcal{N}, \mathcal{E}_s, D)   
\end{equation}
This is comprised of a set $\mathcal{N}$ of $n$ nodes, representing candidate stop locations; a set $\mathcal{E}_s$ of street edges $(i, j, \tau_{ij})$ connecting the nodes, weighted by driving time $\tau_{ij}$; and an $n \times n$ matrix $D$ giving the travel demand (in number of trips) between every pair of nodes in $\mathcal{N}$.  A route $r$ is a sequence of nodes in $\mathcal{N}$.  The goal is to find a set of routes $\mathcal{R}$ that minimize a cost function $C: \mathcal{C}, \mathcal{R} \rightarrow \mathbb{R}^+$.  The route set $\mathcal{R}$ is called a transit network, and is subject to the following constraints:
\begin{enumerate}
    \item $\mathcal{R}$ must allow every node in $\mathcal{N}$ to be reached from every other node via transit.
    \item $\mathcal{R}$ must contain exactly $S$ routes (ie. $|\mathcal{R}| = S$), where $S$ is a parameter set by the user.
    \item $MIN \leq |r| \leq MAX \; \forall \; r \in \mathcal{R}$, where $MIN$ and $MAX$ are parameters set by the user.
    \item A route $r\in\mathcal{R}$ may not contain cycles; each node $i \in \mathcal{N}$ can appear in $r$ at most once.
    \item A route cannot ``skip'' nodes: if nodes $i,j$ are consecutive in $r\in \mathcal{R}$, then $(i, j, \tau_{ij})$ must be in $\mathcal{E}_s$.
\end{enumerate}
We here deal with the symmetric \ac{NDP}, that is: $D = D^\top$,
$(i, j, \tau_{ij}) \in \mathcal{E}_s$ iff. $(j, i, \tau_{ij}) \in \mathcal{E}_s$, and all routes are traversed both forwards and backwards by vehicles on them.

\subsection{Markov Decision Process Formulation}\label{subsec:mdp}

\begin{figure}
    \centering
    \includegraphics[width=0.9\columnwidth]{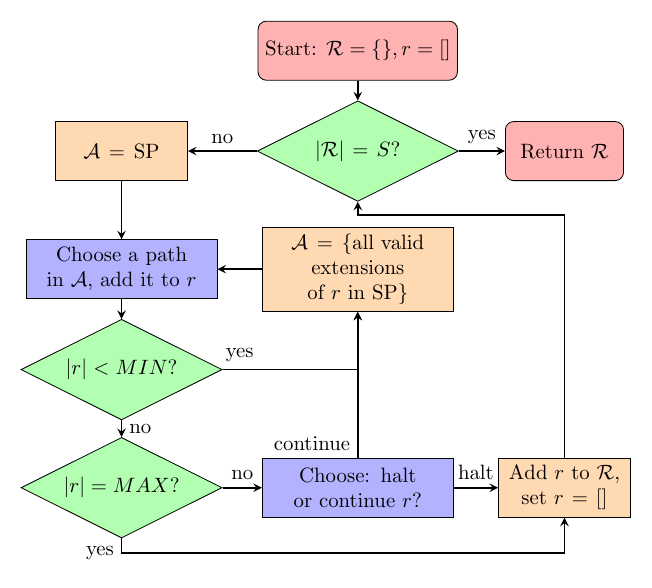}
    \caption{A flowchart of the transit network construction process defined by our \ac{MDP}.  Blue boxes indicate points where the timestep $t$ is incremented and the neural net policy selects an action.}
    \label{fig:mdp_flowchart}
\end{figure}

A \acf{MDP} is a formalism used to define problems in \ac{RL}.  In an \ac{MDP}, an \textbf{agent} interacts with an environment over time steps $t$.  At each $t$, the environment is in \textbf{state} $s_t \in \mathcal{S}$, and the agent takes some \textbf{action} $a_t \in \mathcal{A}_t$, where $\mathcal{A}_t$ is the set of possible actions at $t$.  The environment transitions to a new state $s_{t+1} \in \mathcal{S}$ according to the state transition distribution $P(s' | s, a)$, and the agent receives a scalar \textbf{reward} $R_t \in \mathbb{R}$ according to the reward distribution $P(R | s, a, s')$.  The agent chooses $a_t$ according to its \textbf{policy} $\pi(a|s)$, a probability distribution over $\mathcal{A}_t$ given $s_t$.  In \ac{RL}, the goal is to learn a policy $\pi$ that maximizes cumulative reward.

We here describe the \ac{MDP} we use to represent a construction approach to the \ac{NDP}, visualized in Fig.~\ref{fig:mdp_flowchart}.  The state $s_t$ is composed of the set of routes $\mathcal{R}_t$ planned so far, and an incomplete route $r_t$ which is being planned.  
\begin{equation}\label{eqn:state}
    s_t = (\mathcal{R}_t, r_t)
\end{equation}
The starting state is $s_0 = (\mathcal{R}_0 = \{\}, r_0 = [])$.  The \ac{MDP} alternates with $t$ between two modes: on odd-numbered $t$, the agent chooses an extension to $r_t$; on even-numbered $t$, the agent chooses whether to finish $r_t$ and start a new route.

On odd-numbered $t$, $\mathcal{A}_t$ is drawn from $\textup{SP}$, the set of shortest paths between all node pairs.  If $r_t = []$, then $\mathcal{A}_t = \{a \; | \; a \in \textup{SP}, |a| \leq MAX\}$.  Otherwise, $\mathcal{A}_t$ is comprised of paths $a \in \textup{SP}$ that meet the following conditions:
\begin{itemize}
    \item $(i,j,\tau_{ij}) \in \mathcal{E}_s$, where $i$ is the first node of $a$ and $j$ is the last node of $r_t$, or vice-versa
    \item $a$ and $r_t$ have no nodes in common
    \item $|a| \leq MAX - |r_t|$
\end{itemize}
Once a path $a_t \in \mathcal{A}_t$ is chosen, $r_{t+1}$ is formed by appending $a_t$ to the beginning or end of $r_t$ as appropriate.

On even-numbered $t$, the action space depends on the number of stops in $r_t$:
\begin{align}\label{eqn:halt_actions}
    \mathcal{A}_t = \begin{cases}
        \{\textup{continue}\} & \text{if} |r_t| < MIN \\
        \{\textup{halt}\} & \text{if} |r_t| = MAX \\        
        \{\textup{continue}, \textup{halt}\} & \text{otherwise}
    \end{cases}
\end{align}
If $a_t = \textup{halt}$, $r_t$ is added to $\mathcal{R}_t$ to get $\mathcal{R}_{t+1}$, and $r_{t+1}$  is set to a new empty route.  If $a_t = \textup{continue}$, then $\mathcal{R}_{t+1}$ and $r_{t+1}$ are unchanged from step $t$.


When $|\mathcal{R}_t| = S$, the \ac{MDP} terminates, giving the final reward $R_t = -C(\mathcal{C}, \mathcal{R}_t)$. At all prior steps, $R_t = 0$.

This \ac{MDP} formulation imposes some helpful biases on $\mathcal{R}$.  First, it biases routes towards directness by forcing them to be composed of shortest paths.  Second, the alternation between deciding {\it whether} and {\it how} to extend a route means that the probability of halting does not depend on how many extensions are possible at $t$.

\subsection{Cost Function}

Following earlier work \cite{mumford2013new}, the \ac{NDP} cost function has three components.  The passenger cost is the average transit trip time:
\begin{equation}
    C_p(\mathcal{C}, \mathcal{R}) = \frac{\sum_{i,j} D_{ij}\tau_{\mathcal{R}ij}}{\sum_{i,j} D_{ij}}
\end{equation}
Where $\tau_{\mathcal{R}ij}$ is the time of the shortest transit trip from $i$ to $j$ over $\mathcal{R}$, including a time penalty $p_T$ for each transfer.  

The operator cost is the total driving time of the routes:
\begin{equation}
    C_o(\mathcal{C}, \mathcal{R}) = \sum_{r \in \mathcal{R}} \tau_r
\end{equation}
Where $\tau_r$ is the time needed to completely traverse a route $r$ in both directions.

To enforce the constraints on $\mathcal{R}$, we add a third term $C_c$, which is the fraction of node pairs that are not connected by $\mathcal{R}$ plus a measure of how much $|r| > MAX$ or $|r| < MIN$ across all routes.  The cost function is then:

\begin{equation}
    C(\mathcal{C}, \mathcal{R}) = \alpha w_p C_p + (1 - \alpha) w_o C_o + \beta C_c
\end{equation}

The weight $\alpha \in [0, 1]$ controls the trade-off between passenger and operator costs.  $w_p$ and $w_o$ are re-scaling constants chosen so that $w_p C_p$ and $w_o C_o$ both vary roughly over the range $[0, 1]$ for different $\mathcal{C}$ and $\mathcal{R}$; this is done so that $\alpha$ will properly balance the two, and to stabilize training of the \ac{GNN} policy.  The values used are $w_p = (\max_{i,j}T_{ij})^{-1}$ and $w_o = (3S\max_{i,j}T_{ij})^{-1}$, where $T$ is an $n \times n$ matrix of shortest-path driving times between every node pair.

\section{Methodology}

\subsection{Learned Constructor}

We propose to learn a policy $\pi_\theta(a|s)$ with the objective of maximizing the cumulative return $G$ on the construction \ac{MDP} described in section~\ref{subsec:mdp}.  
By then evaluating this policy on the \ac{MDP} for some city $\mathcal{C}$, we can obtain a transit network $\mathcal{R}$ for that city.  We denote this algorithm the Learned Constructor (LC).  

The policy $\pi_\theta$ is a neural net parameterized by $\theta$.  Its ``backbone'' is a graph attention net \cite{gatv2conv} which treats the city as a fully-connected graph on the nodes $\mathcal{N}$, where each edge has a feature vector $e_{ij}$ containing information about demand, existing transit connections, and the street edge (if one exists) between $i$ and $j$.  We note that a graph attention net operating on a fully-connected graph has close parallels to a Transformer model~\cite{vaswani2017attention}, but unlike a Transformer it can use make use of edge features that describe relationships between nodes.

The backbone \ac{GNN} outputs node embeddings $Y$, which are operated on by one of two policy ``heads'', depending on the timestep: $\textup{NN}_{ext}$ for choosing among extensions when $t$ is odd, and $\textup{NN}_{halt}$  for deciding whether to halt when $t$ is even.  For the full details of the neural architecture, we direct the reader to our code release\footnote[2]{Available at \url{https://www.cim.mcgill.ca/~mrl/hgrepo/transit_learning/}}.

\subsubsection{Training}\label{subsec:methodology_training}

Following the work of~\cite{Kool2019AttentionLT}, we train the policy net using the policy gradient method REINFORCE with baseline~\cite{williams1992reinforce}, setting $\gamma = 1$.  Since the reward $R_t$ for the last step is the negative cost and at all other steps $R_t=0$, this implies the return $G_t$ at each timestep is simply:
\begin{equation}
G_t = \sum_{t'} \gamma^{t' - t} R_t = \sum_{t'} R_t = -C(\mathcal{C}, \mathcal{R})
\end{equation}
The learning signal for each action $a_t$ is $G_t - b(\mathcal{C}, \alpha)$, where the baseline $b(\mathcal{C}, \alpha)$ is a separate \acl{MLP} trained to predict the final reward obtained by the current policy for a given cost weight $\alpha$ and city $\mathcal{C}$.

We train the model on a dataset of synthetic cities.  For each batch, a full rollout of the \ac{MDP} is performed on the cities in the batch, $C(\mathcal{C}, \mathcal{R})$ is computed across the batch, and back-propagation and weight updates are applied to both the policy net and the baseline net.

To construct a synthetic city for the training dataset, we first generate its nodes and street network using one of these processes chosen at random:
\begin{itemize}
    \item $4$-nn: Sample $n$ random points in a square to give $\mathcal{N}$.  Add edges to each node $i$ from its 4 nearest neighbours.
    \item 4-grid: Place $n$ nodes in a rectangular grid as close to square as possible.  Add edges from each node to its horizontal and vertical neighbours.
    \item 8-grid: Like 4-grid, but also add edges between diagonal neighbours.
    \item Voronoi: Sample $m$ random 2D points, and compute their Voronoi diagram~\cite{fortune1995voronoi}.  Take the shared vertices and edges of the resulting Voronoi cells as $\mathcal{N}$ and $\mathcal{E}_s$.  $m$ is chosen so $|\mathcal{N}| = n$.    
\end{itemize}

For each process except Voronoi, each edge in $\mathcal{E}_s$ is then deleted with user-defined probability $\rho$.  If the resulting street graph is not connected, it is discarded and the process is repeated.  Nodes are sampled in a $30 \textup{km} \times 30 \textup{km}$ square, and a fixed vehicle speed of $v = 15 \textup{m/s}$ is assumed to compute street edge weights $\tau_{ij} = ||(x_i, y_i) - (x_j, y_j)||_2 / v$.  Finally, we generate $D$ by setting diagonal demands $D_{ii} = 0$ and uniformly sampling off-diagonal elements $D_{ij}$ in the range $[60, 800]$.


All neural net inputs are normalized so as to have unit variance and zero mean across the entire dataset during training.  The normalization parameters are saved as part of the model and applied to new data presented at test time. 

\subsection{Evolutionary Algorithm}

An \acl{EA} is an improvement method in which an initial population of $B$ solutions go through repeated stages of modification and selection based on their ``fitness'', thus increasing the fitness of the population over many iterations.  In~\cite{nikolic2013transit}, an \ac{EA} is presented for the \ac{NDP}, which consists of alternating mutation and selection stages.  In the mutation stage, two different mutation operators - type 1 and 2 - are each applied $N_m$ times to half of the population.  Each mutation is kept if it decreases the cost $C_b$ of solution $b$.  In the selection stage, each solution $b$ ``dies'' with probability increasing with $C_b$.  Dead solutions are replaced with copies (``offspring'') of surviving solutions to maintain the population size $B$, with probability inversely proportional to the survivors' $C_b$.  These alternating stages repeat for a fixed number of iterations $I$.

Each mutator begins by selecting a random terminal $i$ on a random route $r$ in the solution.  The type-1 mutator then selects a random node $j \neq i$ in $\mathcal{N}$, and replaces $r$ with the shortest path between $i$ and $j$.  The type-2 mutator chooses with probability $0.2$ to delete $i$ from $r$; otherwise, it adds a random node $j$ in $i$'s neighbourhood to $r$ (before $i$ if $i$ is the first node in $r$, and after $i$ if $i$ is the last node in $r$), making $j$ the new terminal.

We modify this algorithm by replacing the type-1 mutator with a ``neural mutator''.  This mutator selects a random route $r$ from $\mathcal{R}$, and then rolls out our learned policy $\pi_\theta$ starting from $\mathcal{R} \setminus r$, generating one new route $r'$ that replaces $r$.  The algorithm is otherwise unchanged.  We replace the type-1 mutator because its action space (replacing one route by a shortest path) is a subset of the action space of the neural mutator (replacing one route by a new route composed of shortest paths), while the type-2 mutator's action space is quite different.  In this way, $\pi_\theta$, which was trained as a construction policy, is used in an improvement method.


\begin{table}
    \caption{Statistics of the five synthetic benchmark cities used in our experiments.}    
    \label{tab:dataset}
    \centering
    \begin{tabular}{lcccccc}
        \toprule
          City & $n$ & $|\mathcal{E}_s|$ & $S$ & $MIN$ & $MAX$ & Area (km$^2$) \\
          \midrule
          Mandl   & 15  & 20  & 6  & 2  & 8 & 352.7 \\
         Mumford0 & 30  & 90  & 12 & 2  & 15 & 354.2 \\
         Mumford1 & 70  & 210 & 15 & 10 & 30 & 858.5 \\
         Mumford2 & 110 & 385 & 56 & 10 & 22 & 1394.3 \\
         Mumford3 & 127 & 425 & 60 & 12 & 25 & 1703.2 \\
         \bottomrule
    \end{tabular}
\end{table}

We note that in \cite{nikolic2013transit}, the algorithm is described as a ``bee colony optimization'' algorithm.  But ``bee colony optimization'' is merely a relabelling of the evolutionary-algorithm metaheuristic - the two are functionally identical~\cite{sorensen2015metaheuristics}.  To avoid the profusion of unnecessary terminology, we here describe~\cite{nikolic2013transit}'s method as an \ac{EA}.

\section{Experiments}
All evaluations are performed on the Mandl~\cite{mandl1980evaluation} and Mumford~\cite{mumford2013dataset} city datasets, two popular benchmarks for evaluating \ac{NDP} algorithms~\cite{mumford2013new, john2014routing, kilic2014demand, lin2022analysis}.  The Mandl dataset is one small synthetic city, while the Mumford dataset consists of four synthetic cities, labelled Mumford0 through Mumford3, and it gives values of $S$, $MIN$, and $MAX$ to use when benchmarking on each city.  The values $n$, $S$, $MIN$, and $MAX$ for Mumford1, Mumford2, and Mumford3 are taken from three different real-world cities and their existing transit networks, giving the dataset a degree of realism.  Details of these benchmarks are given in Table~\ref{tab:dataset}.

In all experiments, we use policies $\pi_\theta$ that we trained on a dataset of $2^{15}$ synthetic cities with $n = 20$.  A 90:10 training:validation split of this dataset is used; after each epoch of training, the model is evaluated on the validation set, and at the end of training, the parameters $\theta$ from the epoch with the best validation-set performance are returned.  Data augmentation is applied each time a city is used for training.  This consists of multiplying the node positions $(x_i, y_i)$ and travel times $\tau_{ij}$ by a random factor $c_s \sim [0.4, 1.6]$, rotating the node positions about their centroid by a random angle $\phi \sim [0^\circ, 360^\circ)$, and multiplying $D$ by a random factor ${c_d \sim [0.8, 1.2]}$.  During training, we use constant values $S=10, MIN=2, MAX=15$; we used these as initial values based on the values for the similarly-sized Mandl and Mumford0 cities, and found they gave good results, so did not vary them.  We sample a different $\alpha \sim [0, 1]$ for each city in a batch, so that the policy will learn to condition its actions on the user preferences implicit in the cost function.  Training proceeds for 5 epochs, with a batch size of 64 cities.  When performing multiple training runs with different random seeds, we hold the dataset constant, but not the data augmentation.

We henceforth refer to \cite{nikolic2013transit}'s evolutionary algorithm as EA, and to our own algorithm as NEA.  For both, we set all algorithmic parameters to the values used in the experiments of~\cite{nikolic2013transit}: population size ${B=10}$, mutations per mutation stage per individual ${N_m=10}$, and total number of mutate-select iterations $I=400$.  We tried running \ac{EA} for up to $I=2,000$ on several cities, but found this did not yield any improvement over $I=400$.  In each mutation stage, the neural mutator is applied to half of the population, and the type-2 mutator to the other half, just as the base \ac{EA} splits type-1 and type-2 mutators equally among individuals.  Hyperparameter settings of the model architecture of $\pi_\theta$ and training process were arrived at by a limited manual search; for their values, we direct the reader to the configuration files contained in our code release.  We set the constraint penalty weight $\beta=5$ in all experiments.

\subsection{Results}

\begin{table*}
    \centering
    \caption{Average final cost $C(\mathcal{C}, \mathcal{R})$ achieved by each method over 10 random seeds, for three different settings of cost weight $\alpha$.  Bold indicates the best in each column.  Orange indicates that one seed's solution violated a constraint, red indicates two or three seeds' solutions did so.  Values after $\pm$ are standard deviations over the 10 seeds.}
\begin{tabular}{ccrllllll}
\toprule
 $\alpha$ & Method & City: &             Mandl &          Mumford0 &          Mumford1 &          Mumford2 &          Mumford3 \\
\midrule

& EA    & & \bf 0.270 $\pm$ 0.048 & \bf 0.272 $\pm$ 0.044 &  0.854 $\pm$ 0.277 &  0.692 $\pm$ 0.277 & \textcolor{orange}{0.853 $\pm$ 0.297} \\
0.0 & LC-100 & &  0.317 $\pm$ 0.076 &  0.487 $\pm$ 0.323 &  0.853 $\pm$ 0.371 &  0.688 $\pm$ 0.180 &  0.710 $\pm$ 0.172 \\
& NEA    & & 0.276 $\pm$ 0.054 & 0.298 $\pm$ 0.122 & \textcolor{orange}{\bf 0.623 $\pm$ 0.163} &  \textcolor{orange}{\bf 0.537 $\pm$ 0.236} & \textcolor{orange}{\bf 0.572 $\pm$ 0.262} \\
 
\midrule
& EA     & &  \bf 0.327 $\pm$ 0.009 &  \bf 0.563 $\pm$ 0.009 & \textcolor{red}{1.015 $\pm$ 0.424} &  0.710 $\pm$ 0.258 &  0.944 $\pm$ 0.274 \\
0.5 & LC-100 & &  0.346 $\pm$ 0.040 &  0.638 $\pm$ 0.138 &  0.742 $\pm$ 0.176 &  0.617 $\pm$ 0.085 &  0.612 $\pm$ 0.082 \\
& NEA    & &  0.331 $\pm$ 0.021 &  0.571 $\pm$ 0.034 & \bf 0.627 $\pm$ 0.049 & \bf 0.532 $\pm$ 0.028 & \bf 0.584 $\pm$ 0.215 \\

\midrule
 & EA & & \bf 0.315 $\pm$ 0.002 &  0.645 $\pm$ 0.030 & \textcolor{orange}{0.739 $\pm$ 0.271} &  0.656 $\pm$ 0.252 &  \textcolor{red}{1.004 $\pm$ 0.401} \\
1.0 & LC-100 & &  0.340 $\pm$ 0.015 &  0.738 $\pm$ 0.042 &  0.600 $\pm$ 0.017 &  0.534 $\pm$ 0.009 &  0.504 $\pm$ 0.007 \\
 & NEA  & &  0.317 $\pm$ 0.004 &  \bf 0.637 $\pm$ 0.036 &  \bf 0.564 $\pm$ 0.018 &  \bf 0.507 $\pm$ 0.007 &  \bf 0.481 $\pm$ 0.007 \\
\bottomrule
\end{tabular}
\label{tab:main}
\end{table*}

We compare LC, \ac{EA}, and NEA on Mandl and the four Mumford cities.  To evaluate LC, we perform 100 rollouts on each city and choose the $\mathcal{R}$ from among them that minimizes $C(\mathcal{C}, \mathcal{R})$ (denoted LC-100).  Each algorithm is run across a range of 10 random seeds, using separate policy parameters $\theta$ trained with that seed.  We report results averaged over all of the seeds.  Our main results are summarized in Table~\ref{tab:main}, which shows results at three different $\alpha$ values, optimizing for the operators' perspective ($\alpha=0.0$), the passengers' perspective ($\alpha=1.0$), and a balance of the two ($\alpha=0.5$).  

The results show that while \ac{EA} performs best on the two smallest cities in most cases, its relative performance worsens considerably when $n \geq 70$.  On Mumford1, 2, and 3, for each $\alpha$, 
NEA performs best overall on these three cities.  It is better than LC-100 in every instance, improving on its cost by about 6\% $\alpha=1.0$ and $0.5$, and by up to 20\% at $\alpha=0.0$; and it improves on \ac{EA} by 33\% to 53\% on Mumford3 depending on $\alpha$.

With $\alpha=0.0$, NEA failed to obey route length limits with 1 of the 10 seeds.  This may be because at $\alpha=0.0$, the reduction in $C_o$ given by under-length routes overwhelms the cost penalty due to a few routes being too long.  
This might be resolved by simply increasing $\beta$ or adjusting the specific form of $C_c$.

We also observed that while one NEA iteration is somewhat slower than one EA iteration due to the need to execute $\pi_\theta$, NEA's cost falls much more rapidly with each iteration than does EA's, so much so that NEA can achieve lower-cost solutions in considerably less time than EA.  This fact may be practically significant: autonomous buses may enable new forms of transit involving online re-planning of bus routes, and in this application, planning high-quality route systems at speed will be key.  NEA may make such an application feasible where EA may not.

\subsection{Trade-offs Between Passenger and Operator Costs}


\begin{figure*}
    \centering
    \begin{subfigure}[b]{0.32\textwidth}
        \centering
        \includegraphics[width=\textwidth]{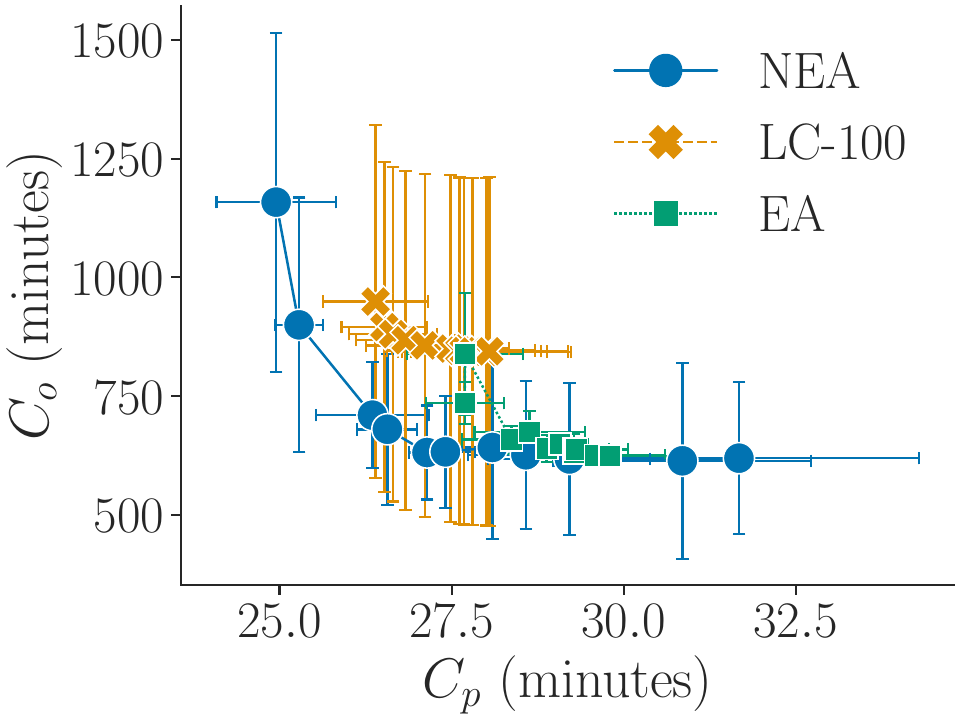}
        \caption{Mumford1}
        \label{subfig:mumford1pareto}
    \end{subfigure}
    \hfill
    \begin{subfigure}[b]{0.32\textwidth}
        \centering
        \includegraphics[width=\textwidth]{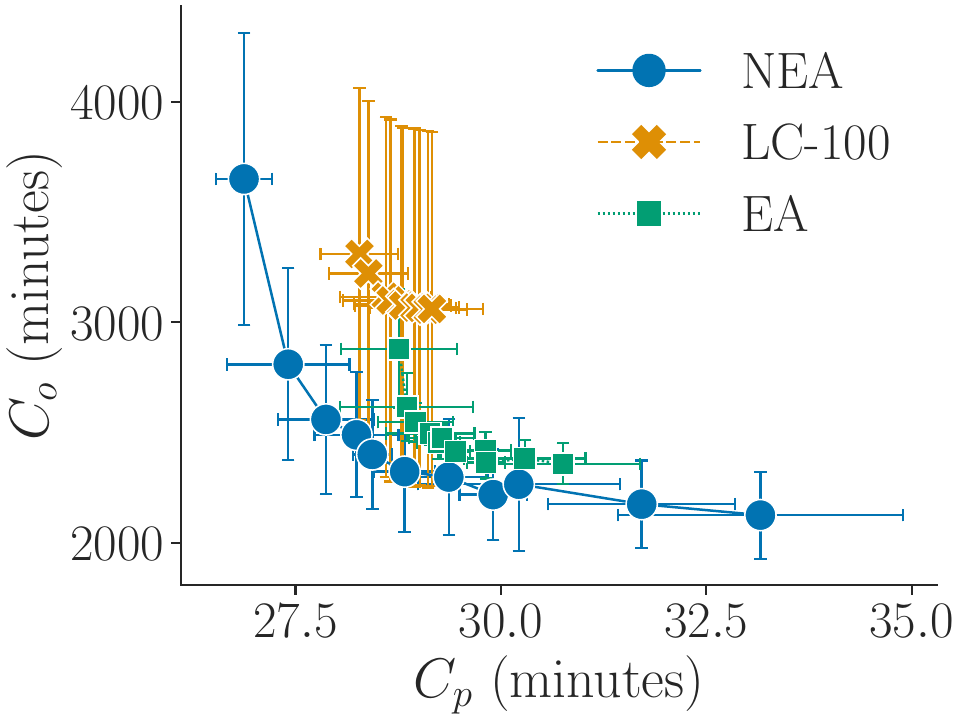}
        \caption{Mumford2}
        \label{subfig:mumford2pareto}
    \end{subfigure}
    \hfill
    \begin{subfigure}[b]{0.32\textwidth}
        \centering
        \includegraphics[width=\textwidth]{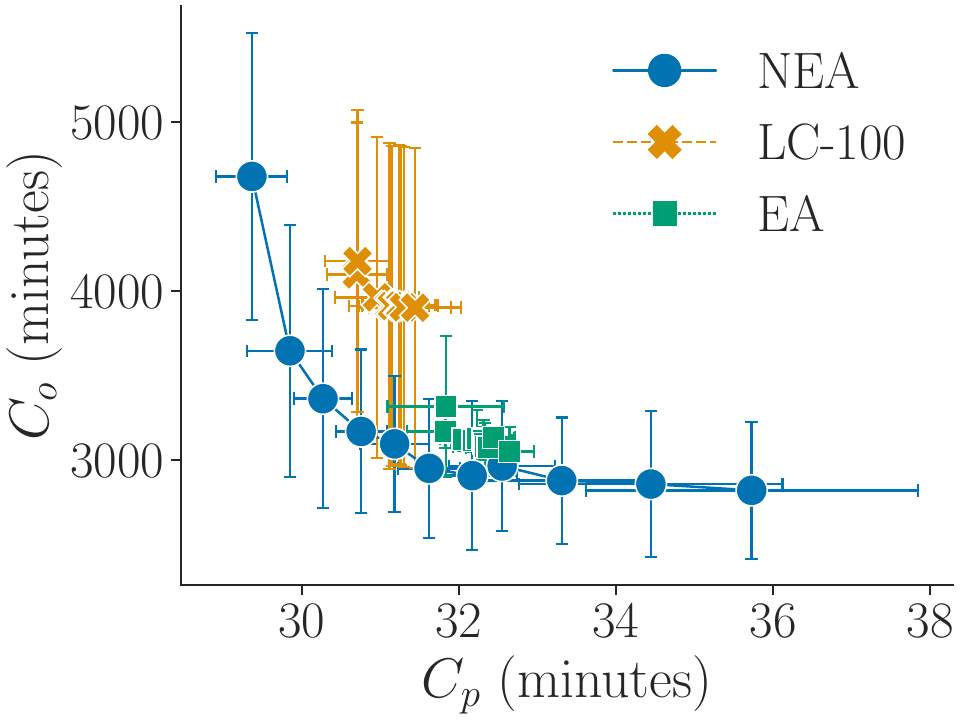}
        \caption{Mumford3}
        \label{subfig:mumford3pareto}
    \end{subfigure}
    \caption{Trade-offs achieved by different methods between passenger cost $C_p$ (on the x-axis) and operator cost $C_o$ (on the y-axis), across values of $\alpha$ evenly spaced over the range $[0, 1]$, averaged over 10 random seeds.  Both axes have units of minutes.  Error bars show one standard deviation over the ten random seeds for each point.  We wish to minimize both values, so the lower-left direction in each plot represents improvement.  
    A line links two points if they have adjacent $\alpha$ values, so these curves show a smooth progression from low $C_o$ (at the right) to low $C_p$ (at the left) as $\alpha$ increases.}
    \label{fig:pareto}
\end{figure*}

There is always a trade-off between minimizing the passenger cost $C_p$ and the operator cost $C_o$: making transit routes longer increases $C_o$, but allows more and faster direct connections between stops, and so may decrease $C_p$.  The weight $\alpha$ can be set by the user to indicate how much they care about $C_p$ versus $C_o$, and each algorithm's behaviour will change accordingly.  Fig.~\ref{fig:pareto} illustrates the trade-offs made by the different methods, as we vary $\alpha$ over the range $[0, 1]$ in steps of $0.1$ on the three cities with real-world statistics, Mumford1, 2, and 3.  NEA's solutions not only dominate those of \ac{EA}, but also achieve a much wider range of $C_p$ and $C_o$ than either of \ac{EA} or LC-100, which will be more satisfactory if the user cares only about one or the other component.  

Both LC and \ac{EA} have more narrow ranges of $C_p$ and $C_o$ on these three cities, but the ranges are mostly non-overlapping.  Some of NEA's greater range seems to be due to combining the non-overlapping ranges of the constituent parts, but NEA's range is greater than the union of LC's and \ac{EA}'s ranges.  This implies that the larger action space of the neural mutator versus the type-1 mutator allows NEA to explore a much wider range of solutions by taking wider ``steps'' in solution space.  

\subsection{Comparison with Other Methods}

\begin{table}
\centering
\caption{Comparison between our method and others from the literature on the three Mumford cities, for $\alpha=0.0$ and $\alpha=1.0$, with values for the corresponding metric.}
\begin{tabular}{cccc}
\toprule
 Method &   Mumford1 &          Mumford2 &          Mumford3 \\
 \midrule
 \multicolumn{4}{c}{$C_o$ at $\alpha = 0.0$}  \\
 \midrule
\cite{mumford2013new} & 568 & 2244 & 2830 \\
\cite{john2014routing} & \bf 462 & \bf 1875 & \bf 2301 \\
NEA & \textcolor{blue}{620 $\pm$ 160} & 2126 $\pm$ 197 & 2819 $\pm$ 404 \\

\midrule
\multicolumn{4}{c}{$C_p$ at $\alpha = 1.0$} \\
\midrule

\cite{mumford2013new} & 24.79 & 28.65 & 31.44 \\
\cite{john2014routing} & 23.91 & 27.02 & 29.50 \\
\cite{kilic2014demand} & \bf 23.25 & \bf 26.82 & 30.41 \\ 
NEA & 24.95 $\pm$ 0.87 & \textcolor{blue}{26.87 $\pm$ 0.34} & \bf 29.37 $\pm$ 0.45 \\
 \bottomrule
\end{tabular}
\label{tab:sota_comparison}
\end{table}

The main purpose of our work was to see whether a neural net could learn route-building heuristics that would improve the results of an existing metaheuristic algorithm, and we have shown that it does.  A direct comparison with other non-neural algorithms from the literature is secondary to this purpose, but is still of some interest.  So Table~\ref{tab:sota_comparison} compares our NEA with more methods from elsewhere in the literature, on the three real-world-based Mumford benchmark cities.  We bold the best result, and show NEA's result in blue if it is within one standard deviation of the best result.  In \cite{mumford2013new, john2014routing}, results are reported for $\alpha=0.0$ (optimizing only $C_o$) and $\alpha=1.0$ (optimizing only $C_p$), while \cite{kilic2014demand} only reports results for $\alpha=1.0$.

At $\alpha=0.0$, we find that NEA performs comparably with \cite{mumford2013new}, but not as well as \cite{john2014routing}.  At $\alpha=1.0$, NEA fares better, outperforming \cite{mumford2013new}, \cite{john2014routing}, and \cite{kilic2014demand} on two of the three cities.  We note that this competitive performance is impressive given the substantial difference in runtime.  While \cite{mumford2013new} do not report run-times with their results, \cite{john2014routing} report that their NSGA-II genetic algorithm takes more than two days to run on Mumford3, while \cite{kilic2014demand} report that their procedure takes eight hours just to construct the initial solution for Mumford3.  By comparison, our NEA runs take less than 20 minutes on a desktop computer with a 2.4 GHz Intel i9-12900F processor and an NVIDIA RTX 3090 GPU.  Running our method with larger values of $I$ or $B$ would increase runtime but may generate better solutions.  Using other, more costly but more powerful metaheuristics with our neural heuristic may also give better results.  We leave this to future work.

\section{Discussion}





We have shown that a neural net policy trained to construct whole solutions to the transit network design problem learns broadly useful heuristics for planning transit routes.  This allows it serve as a mutation operator in an evolutionary algorithm, significantly improving the quality of the discovered solutions on a standard benchmark.  Meanwhile, the evolutionary algorithm's power as a search method enhances results considerably over those achieved by directly sampling solutions from the learned policy.

Better results could likely be achieved by training a policy directly in the context of an improvement process, rather than training it in a construction process as was done here.  It would also be interesting to use an ensemble of separately-trained policies as different heuristics within a \ac{GA}, as opposed to the single model used in our experiments.  Training over variable $n, S, MIN,$ and $MAX$ would be interesting as well.

We note that both the construction policy and the neural evolutionary algorithm outperform the plain evolutionary algorithm on the Mumford1, 2, and 3 problem instances - all those that were modeled on specific real-world cities in scale.  Furthermore, the gap between these grows with the size of the city.  This suggests that our approach may scale better to much larger problem sizes.  This is significant, as many real-world cities have hundreds or thousands of bus stop locations~\cite{montrealTransit}.  In future, we intend to evaluate our technique on data from real-world cities.

Another promising direction would be to extend this work to the problem of planning a transit network that would operate in conjunction with a mobility-on-demand system composed of autonomous taxis.  A fully-autonomous urban transit system will need to make use of both point-to-point autonomous ride-sharing and fixed autonomous mass transit routes to be as useful and efficient as possible~\cite{alonso2018potential, leich2019should}.  The system proposed in this paper could serve as part of an algorithm for planning such a combined autonomous transit system.  Future work could explore how our method could be combined with techniques from the literature on autonomous mobility-on-demand planning~\cite{vakayil2017integrating, ruch2018amodeus, marti2021demand}.


\bibliographystyle{IEEEtran}
\bibliography{IEEEabrv,references}

\begin{acronym}
  \acro{AV}{autonomous vehicle}
  \acro{TSP}{travelling salesman problem}
  \acro{VRP}{vehicle routing problem}  
  \acro{CVRP}{Capacitated \ac{VRP}}
  \acro{NDP}{Transit Network Design Problem}
  \acro{FSP}{Frequency-Setting Problem}
  \acro{DFSP}{Design and Frequency-Setting Problem}
  \acro{SP}{Scheduling Problem}
  \acro{TP}{Timetabling Problem}
  \acro{NDSP}{Network Design and Scheduling Problem}
  \acro{HH}{hyperheuristic}
  \acro{GA}{genetic algorithm}
  \acro{EA}{evolutionary algorithm}
  \acro{SA}{Simulated Annealing}
  \acro{ACO}{ant colony optimization}
  
  \acro{MoD}{Mobility on Demand}
  \acro{AMoD}{Autonomous Mobility on Demand}
  \acro{IMoDP}{Intermodal Mobility-on-Demand Problem}
  \acro{OD}{Origin-Destination}
  \acro{CSA}{Connection Scan Algorithm}

  \acro{CO}{combinatorial optimization}
  \acro{NN}{neural net}
  \acro{RNN}{recurrent neural net}
  \acro{ML}{Machine Learning}
  \acro{MLP}{Multi-Layer Perceptron}
  \acro{RL}{reinforcement learning}
  \acro{DRL}{deep reinforcement learning}
  \acro{GNN}{graph neural net}
  \acro{MDP}{Markov Decision Process}
  \acro{DQN}{Deep Q-Networks}
  \acro{ACER}{Actor-Critic with Experience Replay}
  \acro{PPO}{Proximal Policy Optimization}
  \acro{ARTM}{Metropolitan Regional Transportation Authority}
\end{acronym}

\end{document}